\documentclass[twoside,11pt]{article}
\usepackage{todonotes}
\usepackage{framed}
\usepackage[preprint]{jmlr2e}

\usepackage{jmlr2e}
\usepackage{listings}

\jmlrheading{1}{2020}{1-48}{4/00}{10/00}{meila00a}{Lucca Portes Cavalheiro, Jean Paul Barddal, Alceu de Souza Britto Junior, Laurent Heutte}

\ShortHeadings{scikit-dyn2sel - A Dynamic Selection Framework for Data Streams}{Cavalheiro, Barddal, Britto Jr., and Heutte}
\firstpageno{1}

\begin{document}

\title{\texttt{scikit-dyn2sel} - A Dynamic Selection Framework for Data Streams}
\author{\name Lucca Portes Cavalheiro \email lucca@ppgia.pucpr.br \\
       \name Jean Paul Barddal \email jean.barddal@ppgia.pucpr.br \\
       \name Alceu de Souza Britto Jr. \email alceu@ppgia.pucpr.br \\
       \addr Graduate Program in Informatics (PPGIa)\\
       Pontif\'{i}cia Universidade Cat\'{o}lica do Paran\'{a} (PUCPR)\\
       Curitiba, Brazil
       \AND
       \name Laurent Heutte \email laurent.heutte@univ-rouen.fr \\
       \addr Laboratoire d'Informatique, du Traitement de l'Information et des Systèmes (LITIS)\\
       Universit\'{e} de Rouen Normandie\\
       Rouen, France}

\editor{?}

\maketitle

\begin{abstract}
Mining data streams is a challenge per se. It must be ready to deal with an enormous amount of data and with problems not present in batch machine learning, such as concept drift. Therefore, applying a batch-designed technique, such as dynamic selection of classifiers (DCS) also presents a challenge. The dynamic characteristic of ensembles that deal with streams presents barriers to the application of traditional DCS techniques in such classifiers. \texttt{scikit-dyn2sel} is an open-source python library tailored for dynamic selection techniques in streaming data. \texttt{scikit-dyn2sel}'s development follows code quality and testing standards, including PEP8 compliance and automated high test coverage using \texttt{codecov.io} and \texttt{circleci.com}. Source code, documentation, and examples are made available on GitHub at \url{https://github.com/luccaportes/Scikit-DYN2SEL}.
\end{abstract}


\section{Introduction}

Dynamic selection of classifiers (DCS) is a widely studied area in batch machine learning. Its application provided significant gains in many types of data. When developing new DCS techniques, it is essential to compare this novel method to the current state-of-art of the area. Nowadays, this is a straightforward task, thanks to \texttt{deslib} \citep{cruz:2020}, a library that allows the application of most DCS methods following a familiar and straightforward interface, borrowed from \texttt{scikit-learn} \citep{pedregosa:2011}.

When dealing with DCS in data stream mining, however, there is no such convenience. 
The application of data stream mining differs from batch machine learning, and thus, it is impossible to apply traditional DCS techniques as is. 
Common concepts in DCS are not naturally present in the streaming environment, such as the validation set, which makes the utilization of \texttt{DESLIB} \citep{cruz:2020} not directly possible. 
In this paper, we propose \texttt{scikit-dyn2sel}, a framework for using and implementing DCS techniques in the data stream mining context.

\section{Structure}
The \texttt{scikit-dyn2sel} framework is built on top of \texttt{scikit-multiflow} \citep{montiel:2018} and \texttt{deslib} \citep{cruz:2020}, a \texttt{scikit-learn} \citep{pedregosa:2011} inspired library for data stream mining. The interface of all the methods for applying DCS follows the same interface as \texttt{scikit-multiflow} classifiers, the essential methods are \texttt{partial\_fit} and \texttt{predict}. These methods are respectively used for updating the classifiers with new data and for computing predictions.

The framework is divided into four main classes. One of these is the \texttt{DCSTechnique} class, which contains the traditional DCS methods implemented. 
The objective of this class is to output a prediction using an ensemble and a validation set, such that the latter is defined in the \texttt{ValidationSet} class.
Some methods for applying DCS can be used directly on traditional online ensembles; however, many also contemplate the ensemble construction step, that is why each method inherits its ensemble from \texttt{Ensemble}. 
All of these classes are combined in the \texttt{ApplyDCS} class, which is the class that the methods for applying DCS in data streams inherit from. This class follows the same interface as \texttt{scikit-multiflow} \citep{montiel:2018}.

Another benefit from \texttt{scikit-dyn2sel} is that traditional DCS techniques available on \texttt{DESLIB} \citep{cruz:2020} are not re-implemented.
Instead, they encapsulated on the \texttt{DCSTechniques} class.

\subsection{Implemented Methods}
Table \ref{tab:dyn2sel} presents all the DCS methods currently implemented in \texttt{scikit-dyn2sel}. 
The left part of the table displays the methods for applying dynamic selection techniques in data streams, and the right part displays the techniques itself.

\begin{table}[]
    \centering
    \caption{Methods Contemplated in \texttt{scikit-dyn2sel}.}
    \resizebox{0.85\textwidth}{!} {%
    \begin{tabular}{||c c | c c||}
        \hline
        DCSApply & DCSTechnique \\ [0.5ex] 
        \hline\hline
        DYNSE \citep{almeida:2016}  & KNORA-E \citep{alceu:2008} \\ 
        \hline 
        DESDD \citep{albuquerque:2019} & KNORA-U \citep{alceu:2008} \\ 
        \hline 
        MDE \citep{wozniak:2019} & A Priori and A Posteriori \citep{Giacinto:1999} \\ 
        \hline
        & DCS-LA \citep{kegelmeyer:1996} \\ 
        \hline
        & DCS-RANK \citep{sabourin:1993} \\ 
        \hline
        & KNOP \citep{cavalin:2013} \\ 
        \hline
         & MCB \citep{huang:1995} \\ 
        \hline
         & META-DES \citep{cruz:2015} \\ 
        \hline
    \end{tabular}
    }
    \label{tab:dyn2sel}
\end{table}

\section{Open Source}
\texttt{scikit-dyn2sel} is open to contributions from the community. It is hosted in a public repository on Github. It is licensed under the MIT license, which is a very embracing and permissive licensing, allowing but not limited to commercial use, distribution, modification, and private use.

\section{Installation}
The installation of the library can be done via Python package manager (\texttt{pip}) using ``\texttt{pip install scikit-dyn2sel}'', or by directly cloning its GitHub repository.

\section{Tests}
To ensure the good operation of the framework, unit tests were written for each main method in the library. When a new contribution to the code is proposed, a continuous integration tool (CircleCi) runs the tests to ensure that if the contribution is accepted, the previously expected behavior of the methods is still respected. To measure the percentage of test coverage, Codecov is applied after CircleCi's tests pass. A contribution is only accepted if it does not decrease the test coverage percentage of the framework.

\section{Code Quality}
The code is fully compliant with Python PEP8 standards, which is ensured by the Black code formatting tool \citep{black}, which is also run on CircleCi after each contribution proposal. Furthermore, the static analyzer Codacy is also integrated into the Github repository, ensuring standardized code quality.

\section{Usage}
The usage of \texttt{scikit-dyn2sel} is straightforward. Since it follows the same interface as \texttt{scikit-multiflow} \citep{montiel:2018}, the methods can be executed with common evaluator used in the library, such as prequential \citep{gama:2013}. 
Figure \ref{fig:code_example} shows how this can be done using the \texttt{DYNSE} \citep{almeida:2016} method.

\begin{figure}
    \centering
    \begin{lstlisting}[language=Python, linewidth=\textwidth, frame = single]
from skmultiflow.evaluation import EvaluatePrequential
from skmultiflow.data import SEAGenerator
from dyn2sel.apply_dcs import DYNSEMethod
from dyn2sel.dcs_techniques import KNORAE

clf = DYNSEMethod(
    HoeffdingTree(), chunk_size=1000, 
    dcs_method=KNORAE(), max_ensemble_size=10)
gen = SEAGenerator()
ev = EvaluatePrequential()
ev.evaluate(gen, clf)
    \end{lstlisting}
    \caption{Usage example of \texttt{scikit-dyn2sel}.}
    \label{fig:code_example}
\end{figure}


\acks{This study was financed in part by the Coordenação de Aperfeiçoamento de Pessoal de Nível Superior - Brasil (CAPES) - Finance Code 001.}

\vskip 0.2in
\bibliography{references}

\end{document}